\documentclass[10pt, a4paper]{article}

\usepackage{lrec-coling2024} 

\usepackage{algorithm}
\usepackage{algorithmic}
\usepackage{multirow}
\usepackage{amsmath}

\title{\textbf{TARN-VIST: Topic Aware Reinforcement Network for \\ Visual Storytelling}}

\name{Weiran Chen, Xin Li, Jiaqi Su, Guiqian Zhu, Ying Li, Yi Ji$^{\ast}$, Chunping Liu$^{\ast}$ \thanks{*Corresponding author}} 

\address{School of Computer Science and Technology, Soochow University, Suzhou, China \\
\{wrchen2023, 20224227052\}@stu.suda.edu.cn, czu\_lixin@163.com \\
\{gqzhu, ingli, jiyi, cpliu\}@suda.edu.cn \\}

\abstract{As a cross-modal task, visual storytelling aims to generate a story for an ordered image sequence automatically. Different from the image captioning task, visual storytelling requires not only modeling the relationships between objects in the image but also mining the connections between adjacent images. Recent approaches primarily utilize either end-to-end frameworks or multi-stage frameworks to generate relevant stories, but they usually overlook latent topic information. In this paper, in order to generate a more coherent and relevant story, we propose a novel method, \textbf{T}opic \textbf{A}ware \textbf{R}einforcement \textbf{N}etwork for \textbf{VI}sual \textbf{S}tory\textbf{T}elling (\textbf{TARN-VIST}). In particular, we pre-extracted the topic information of stories from both visual and linguistic perspectives. Then we apply two topic-consistent reinforcement learning rewards to identify the discrepancy between the generated story and the human-labeled story so as to refine the whole generation process. Extensive experimental results on the VIST dataset and human evaluation demonstrate that our proposed model outperforms most of the competitive models across multiple evaluation metrics.
 \\ \newline \Keywords{Visual Storytelling, Topic Information, Reinforcement Learning} }

\begin{document}

\maketitleabstract

\section{Introduction}

Nowadays, visual storytelling has garnered increasing attention from the fields of both computer vision (CV) and natural language processing (NLP) due to its significance and practicality in some applications such as image retrieval, image subtitling, and blind navigation~\cite{1}. As opposed to visual captioning, visual storytelling also involves exploring the corresponding relationships between object pairs in adjacent images. Additionally, when humans tell stories, they usually revolve around a specific central topic. Therefore, to generate high-quality stories, visual storytelling models also require taking the topic information into account.

In the field of visual storytelling, existing methods can be divided into two main categories: end-to-end-based methods \cite{29,32,3,42,4,2,6,14,31,38,40} and multi-stage-based methods \cite{8,7,9}. Both them, end-to-end-based methods typically map directly from the input image sequence to the output story, while multi-stage-based methods employ different modules trained independently in distinct stages, and the results of the previous stage are always used as the input for the subsequent one. The common idea behind the end-to-end-based methods is to use a convolutional neural network (CNN) as an encoder to extract high-dimensional image features and overall image-stream features. These representational feature vectors are then fed into a long short term memory (LSTM) to construct the story. These approaches can always yield outstanding stories with high score in automatic metrics. On the other hand, multi-stage-based methods, known as planning and writing strategies, advocate for separating the generation process into several steps. Generally speaking, the first step is always to employ an object detection module to detect salient concepts in the given images and the next step is to generate a related story through a transformer-based architecture. This kind of methods can generate stories that reflect human preferences.

Despite their remarkable progress, there are still several technical limitations in the visual storytelling task. One of the drawbacks is that few models consider the latent topic information of the generated story. The topic of the story serves as its central idea, which is the core of the story. It is well known that maintaining a coherent and engaging storyline is crucial, and centering the narrative around a specific topic aids in achieving this coherence. Besides, telling a story around a concrete topic can enhance readers' comprehension as well as facilitate better content retention. Nevertheless, once the generated story goes off-topic, it will become incoherent and lack narrative variety. To be noted, reinforcement learning provides a unique technique to guide the model towards generating stories that not only capture the correct relationships between images but also ensure a  more consistent and accurate topic information. Hence, we incorporate reinforcement learning into our methodology to enhance the overall quality of visual storytelling.

\begin{figure*}[htbp]
\centering
\includegraphics[width=\textwidth, height=0.475\textwidth]{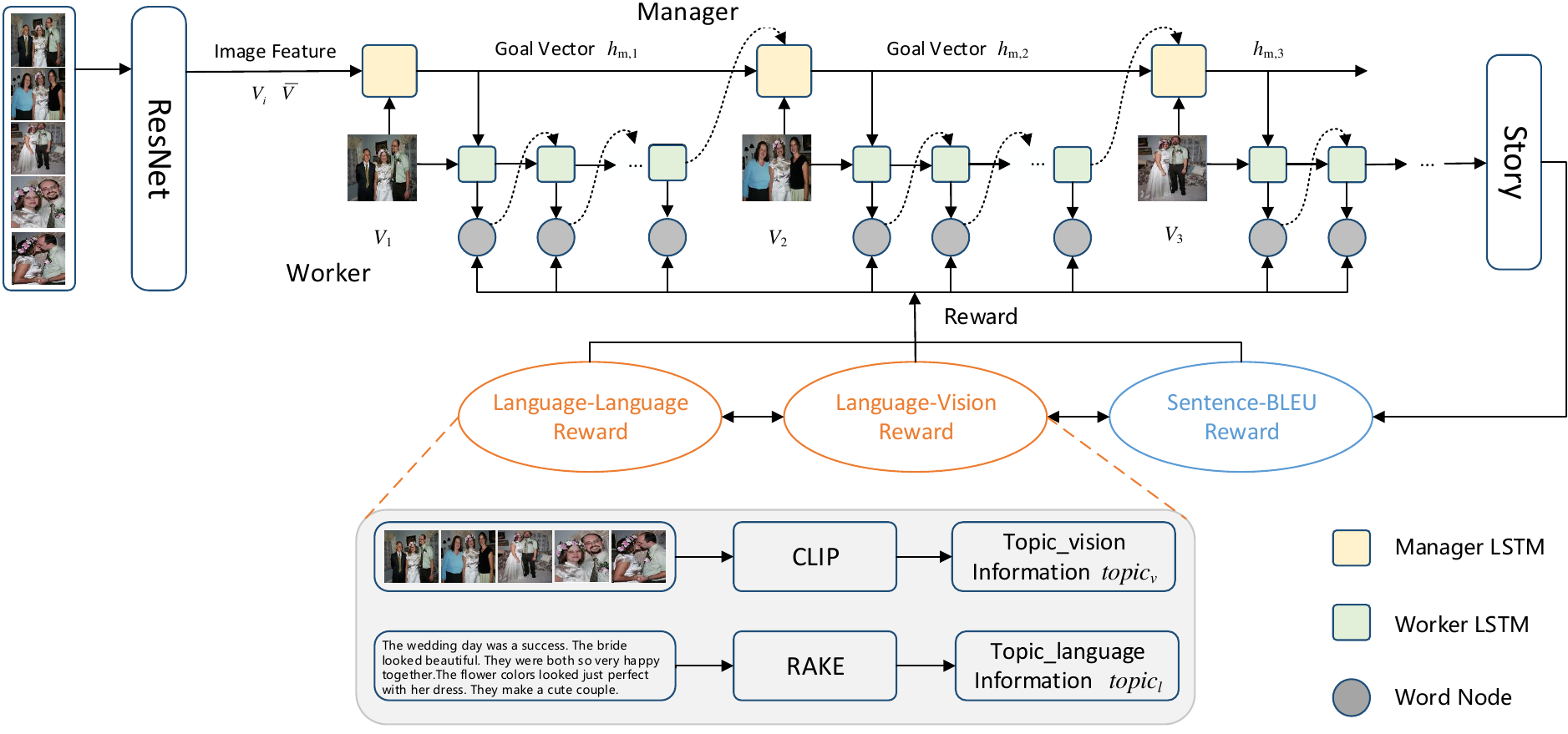}
\caption{Overview of TARN-VIST. In our model, image features are obtained by the pre-trained ResNet and then fed into the hierarchical decoder which consists of a manager LSTM and a worker LSTM to generate a sample story. Once the candidate story is generated, the two topic consistency rewards are combined to refine the generation process. Furthermore, we also set a classical sentence-level BLEU reward to control the fluency of the generated story.}
\label{Fig1}
\end{figure*}

Inspired by the above idea, we propose a novel method called \textbf{T}opic \textbf{A}ware \textbf{R}einforcement \textbf{N}etwork for \textbf{VI}sual \textbf{S}tory\textbf{T}elling (\textbf{TARN-VIST}). As depicted in Fig.~\eqref{Fig1}, our model is an encoder-decoder architecture based on reinforcement learning. To be specific, we first use contrastive language-image pre-training (CLIP) and rapid automatic keyword extraction (RAKE) to extract topic information of the stories in the dataset from both visual and linguistic perspectives, respectively. Subsequently, we harness the extracted topic information along with cosine similarity to design the topic-consistent reinforcement learning rewards. The entire framework aggregates the aforementioned rewards and employs the reinforcement learning algorithm to optimize them.

To summarize, our contributions in this paper are as follows:

\begin{itemize}
\item We first take advantage of CLIP and RAKE together to extract topic information of stories from both visual and linguistic perspectives.
\item To make full use of the topic information, we design reinforcement learning rewards for topic consistency based on the extracted topic information and cosine similarity.
\item  Experimental results on the VIST dataset and human evaluation demonstrate that our proposed model outperforms most of the leading models on multiple evaluation metrics.
\end{itemize}

\section{Related Work}
\subsection{Visual Storytelling}

Visual storytelling is the task of generating a reasonable paragraph-level story with the image sequence selected from a photo stream as input. It necessitates a deeper understanding of the event progression in the image stream. Based on their technical characteristics, we can classify the current visual storytelling techniques into end-to-end frameworks and multi-stage frameworks. Where end-to-end frameworks mainly make use of CNN structure such as VGG~\cite{10} or ResNet~\cite{11} to extract visual features and then apply LSTM to generate story. For instance,~\citet{29} encode an image sequence by running a recurrent neural network (RNN) over the fc7 vectors of each image and use gated recurrent units (GRUs) for both the image encoder and story decoder.~\citet{32} design a model composed of three hierarchically-attentive RNNs to encode the album photos, select representative photos, and compose the story.~\citet{3} put forward a deep learning network model called GLAC net that combine global-local (glocal) attention and context cascading mechanisms together to generate story.~\citet{4} employ a scene encoder and a photo encoder to detect the scene changes and meanwhile aggregate the scene information.~\citet{12} present a commonsense-driven generative model, which introduces related crucial commonsense from the external knowledge base.~\citet{13} raise a novel graph-based architecture for visual storytelling by modeling the two-level relationships on scene graphs.~\citet{30} propose to explicitly learn to imagine the storyline that bridges the visual gap.~\citet{7} introduce two novel modules that consider both the correlation among candidate concepts and the image-concept correlation.~\citet{37} present a novel Latent Memory-augmented Graph Transformer (LMGT), which directly inherits the merits from the Transformer structure.~\citet{6} put forward a novel imagine-reason-write generation framework (IRW), which employs a relational reasoning module to fully exploit the external knowledge and task-specific knowledge.~\citet{14} develop a novel message-passing-like algorithm for ordered image attention (OIA) that collects interactions across all the images in the sequence.~\citet{38} propose a multi-tasking memory-augmented framework, which is jointly trained on factual visual storytelling data and unpaired style corpus.~\citet{41} put forward a coherent visual storytelling (CoVS) framework. It introduces an image sequence encoder module and a new parallel top-down attention module. This kind of approaches can handle the mapping relationship between images and text information well. Their model structures are relatively easy and simple to deploy. However, such methods lack the capacity to model the global structure, and the generated stories usually lack diversity. 

In comparison to end-to-end frameworks, multi-stage frameworks customarily separate the training process into multiple steps and the outputs of the previous step are often used as the input of the subsequent step. For example,~\citet{8} put forward a distill-enrich-generate framework. It distills a set of representative words from the input prompts and enriches the word set by using external knowledge graphs.~\citet{9} introduce PR-VIST to represent the input image sequence as a story graph in which it finds the best path to form a storyline.~\citet{31} propose a new sentiment-aware generative model for VIST called SentiStory. It uses a multi-layered sentiment extraction module (MLSEM). Besides, some researchers have attempted to extract the topic information of stories and generate higher-quality content that closely related to the topic information.~\citet{15} directly use the query vocabulary of the dataset itself as the topic information of the story, but since the query vocabulary is relatively extensive and a topic vocabulary sometimes corresponds to hundreds of sample stories, it cannot accurately represent the theme of the story vocabulary. In addition, visual storytelling is essentially a multi-modal task where the input image sequence also contains plentiful information, but previous studies do not consider the topic vocabulary from the visual perspective.

\subsection{Reinforcement Learning}

Reinforcement learning is an important branch of machine learning where an agent learns how to make optimal decisions and maximize the return of rewards obtained by interactions with a complicated environment~\cite{33,34}. The core concept of reinforcement learning is to learn the most appropriate policy directly through trial and error experiences. The agent performs an action in the environment (Action), and then observes the feedback of the environment (Reward or Punishment), and modifies its behavior in response to the feedback, so as to increase the chances of receiving better rewards in the future~\cite{17}. In recent years, reinforcement learning has been widely used in numerous fields, such as autonomous driving~\cite{16}, robot control~\cite{35}, gaming~\cite{43} and healthcare~\cite{36}, etc. 

For the visual storytelling task, some researchers have already tried to introduce reinforcement learning into the field of visual stories and achieved promising generation results.~\citet{5} propose an adversarial reward learning (AREL) framework to learn an implicit reward function from human demonstrations, and then optimize policy search with the learned reward function.~\citet{42} design the rewards with two critic networks, including a multi-modal and a language-style discriminator to generate relevant and story-style paragraphs. Furthermore, through rethinking about principles that make up high-quality story, \citet{2} present three assessment criteria which are relevance, coherence and expressiveness, and then employed a reinforcement learning framework called ReCo-RL to capture the essence of these quality criteria. However, to the best of our knowledge, attempts on formulating reward functions for reinforcement learning based on topic information are still in the blank stage. Hence, we first extract the topic information of the dataset and design topic-consistent reinforcement learning reward functions to improve the overall generation process.

\section{Approach}

In this paper, we define the visual storytelling task as follows: given an image sequence $I=\left\{i_{1}, i_{2},\ldots, i_{m}\right\}$, the task aims to produce a human-like story $S=\left\{s_{1}, s_{2}, \ldots, s_{m}\right\}$ where $s_{i}$ is a sequence of words describing the $i$-th image.

\subsection{Topic Information Extraction}

In this section, we first describe the topic information extraction process. As the visual storytelling task is a multi-modal task, we utilize CLIP~\cite{18} and RAKE~\cite{19} to extract the topic information of the story from the visual and linguistic perspectives separately.

We use CLIP to extract the topic information of the story from the visual perspective as shown in algorithm \ref{Algo1}. The algorithm mainly includes four steps: candidate-concept extraction, image encoding, text encoding and similarity calculation. 

In the candidate concept extraction step, the $Clarifai's$ image detection API\footnote{https://clarifai.com/clarifai/main/models} is applied to retrieve the concepts that appear in the input image sequence. Different from other common object detection algorithms, this API can not only accurately detect the objects, but also effectively identify the scene and text information in the image stream. At this point, it should be noticed that for each picture, only the top three concepts are taken. So we need to remove some noise concepts (such as "people", "person", "men", etc.). Next, we use the sentence pattern "The topic of this photo is \{concept\}" to assemble the extracted \{concept\} into the sentence pattern. For example, for the concept "graduation", the sentence formed is "The topic of this photo is graduation". The reason for assembling concepts into sentences is to use CLIP for zero-shot prediction, and the pre-training of CLIP is performed on a large number of "sentence-picture" collections. Apart from this, after assembling into sentences, the text encoder can acquire richer text feature information, which is helpful and beneficial for subsequent model processing.

\begin{algorithm}[tb]
	\caption{Visual Perspective Topic Information Extraction Process}
	\label{Algo1}
	\textbf{Input}: Image Sequence $I$  with $5$ images and $Candidate-Concept$\\
	\textbf{Output}: Topic Information \\
	\vspace{-5mm}
	\begin{algorithmic}[1] 
            \STATE Initialise $Candidate-Concept \gets [\ ]$
            \STATE /* Candidate-Concept Extraction */
		\FOR{$i=1$ to $5$} \do \\
		\STATE Extract top-$3$ concept $c_{i}^{j}$ from Image $i$ with $Clarifai's$ Image Detection API
            \STATE Filter out some useless concepts and assemble concepts into sentences $s_{i}^{j}$
        \STATE $Candidate-Concept$. append ($s_{i}^{j}$)
		\ENDFOR 
        \STATE /*                    
      Image Encoding                      */
        \FOR{$i=1$ to $5$} \do \\
        \STATE Image-feature $i$ = CLIP-Image-Encoder (Image $i$)
        \ENDFOR
        \STATE Image-mean-feature = mean (Image-feature $i$) \\
        \STATE /*     Text Encoding     */
        \STATE Text-features=CLIP-Text-Encoder($Candidate$-$Concept$)
        \STATE /*     Similarity Calculation     */
        \STATE Similarity = Image-mean-feature @ Text-features
        \STATE Topic Information = Similarity[0]. topk(1)
        \STATE return Topic Information
	\end{algorithmic}
\end{algorithm}

For the image encoding and text encoding processes, we primarily leverage CLIP's text encoder, image encoder, and zero-sample prediction function. The zero-shot prediction function is realized through the semantic knowledge learned in the pre-training phase, and it can perform image and text classification without the need for additional sample data. Besides, CLIP can comprehend image content through text annotation and image classification. At this stage, the text features of the above text information and the visual features of the input image are extracted by using the text encoder and the image encoder of CLIP, respectively. The overall image features are then calculated by average weighting and adding the individual image features. Finally, we take advantage of cosine similarity to calculate the information with the highest similarity between text features and image features, and then extract the top 1 element from the tensor Similarity[0]. It is worth noting that the topk function is commonly used to retrieve the top k elements from a tensor in practice and the eventual extracted element is a word or phrase, which is the desired image topic information.

Furthermore, we also apply RAKE to extract the topic vocabulary of the stories as the topic information from the linguistic perspective in the dataset. Compared to latent dirichlet allocation (LDA), RAKE can not only rapidly extract keywords from extensive textual data but also excels in identify and extract multi-word phrases, rather than limited to some single keywords. It is versatile, effective, fast and applicable for various text types, including both long and short texts. The core idea behind RAKE is to identify words or phrases with high frequency and importance in the text as the topic information.

\subsection{Topic Aware Storytelling Model}

\textbf{The Encoder} \ We first feed the image sequences into ResNet \cite{11} and extract their high-level image features. For the follow-up requirements, we compute the average value of all the image features ${v}_{i}$ as follows: 

\begin{equation}
\label{eq1}
\overline{\mathrm{V}}=\frac{1}{n} \sum_{i} \mathrm{v}_{i}
\end{equation}

Then, the image features ${v}_{i}$ and overall image sequences features $\overline{\mathrm{V}}$ are combined by direct concatenation along the channel dimension and sent into decoder together.

\textbf{The Decoder} \ The decoder is hierarchical which involves a manager LSTM and a worker LSTM \cite{20}. The manager LSTM serves as a supervisor to control the overall flow of the story, which is denoted as follows: 

\begin{equation}
\label{eq2}
\mathrm{h}_{m, i}=\operatorname{LSTM}_{M}\left(\left[\overline{\mathrm{V}} ; \mathrm{v}_{i} ; \mathrm{h}_{w, i-1}^{T}\right], \mathrm{h}_{m, i-1}\right)
\end{equation}

When describing $i$-th image, the manager LSTM will take three kinds of information into consideration, which are: 1) overall information of the image sequence $\overline{\mathrm{V}}$; 2) image information in the $i$-th image $\mathrm{v}_{i}$; 3) sentences generated for previous image $\mathrm{h}_{w, i-1}^{T}$. Then, the manager LSTM will predict a hidden state $h_{m,i}$ as the goal vector and passes the vector to the worker LSTM.

The worker LSTM attempts to complete the generation of word description based on the goal vector. When generating $t$-th word for the $i$-th image, the worker LSTM decoder will take the current image information $v_{i}$, word embedding of the previously generated word $e_{i}^{t-1} $ and goal vector $h_{m,i}$ as input to predict its hidden state ${h}_{w, i}^{t}$. Then, the worker LSTM enforces a linear layer $f(\cdot)$ to approximate the probability of choosing the next word: 

\begin{equation}
\label{eq3}
\mathrm{h}_{w, i}^{t}=\operatorname{LSTM}\left(\left[v_{i};h_{m,i}; e_{i}^{\mathrm{t}-1}\right], h_{w, i}^{t-1}\right)
\end{equation}

\begin{equation}
\label{eq4}
p_{\theta}\left(y_{i}^{t} \mid y_{i}^{1: t-1}, v_{i}, \bar{V}\right)=\operatorname{softmax}\left(f\left(h_{w, i}^{t}\right)\right)
\end{equation}

\begin{table*}[t]
  \centering
    \begin{tabular}{c|ccccccc}
    \hline
    Method & BLEU-1 & BLEU-2 & BLEU-4 & METEOR & ROUGE & CIDEr & SPICE \\
    \hline
    Seq2seq & - & - & 3.5 & 31.4 & - & 6.8 & - \\
    H-Attn-Rank  & - & - &- & 34.1 & 29.5 & 7.5 & - \\
    BARNN & - & - & - & 33.3 & - & - & - \\
    SRT & 43.4 & 21.4 & 5.2 & 12.3 & - & 11.4 & - \\
    XE-ss  & 62.3  & 38.2  & 13.7  & 34.8  & 29.7  & 8.7   & - \\
    AREL  & 63.7  & 39.0  & 14.0  & 35.0  & 29.6  & 9.5   & 8.9  \\
    HPSR  & 61.9  & 37.8  & 12.2  & 34.4  & \textbf{31.2}  & 8.0   & - \\
    HSRL  & -     & -     & 12.3  & 35.2  & 30.8  & 10.7  & 7.5  \\
    SGVST  & 65.1  & 40.1  & 14.7  & \textbf{35.8} & 29.9  & 9.8   & - \\
    ReCo-RL  & -     & -     & 12.4  & 33.9  & 29.9  & 8.6   & 8.3  \\
    INet  & 64.4  & 40.1  & 14.7  & 35.6  & 29.6  & 11.0  & - \\
    IRW  & 66.7  & 41.6  & \textbf{15.4} & 35.6  & 29.6  & 11.0  & - \\
    CKAKS  & -     & -     & 12.0  & 35.4  & 30.0  & 10.5  & - \\
    LGMT  & 67.5 & 41.6 & 15.1 & 35.6 & 29.7 & 10.0 & - \\
    Sentistory  & 64.8  & 39.8  & 14.2  & 35.3  & 29.8  & 9.7   & - \\
    \hline
    TARN-VIST  & \textbf{69.0} & \textbf{43.5} & 13.4  & \textbf{35.8} & 29.5  & \textbf{12.1} & \textbf{11.3 } \\
    \hline
    \end{tabular}%
  \caption{\label{topic-vist-tabel1} Quantitative results on the VIST dataset for surface-level-based  automatic metrics. For all these metrics, higher score means better performance.}%
\end{table*}%

\begin{table*}[t]
  \centering
    \begin{tabular}{c|ccc}
    \hline
    Method & BERTScore & BARTScore & BLEURT \\
    \hline
    KE-VIST(No KG)  & 28.25 & 17.21 & 43.63 \\
    KE-VIST (With OpenIE) & 29.12 & 17.93 & 46.85 \\
    KE-VIST (with VG)  & 29.16 & 18.03 & 47.54 \\
    PR-VIST  & 27.64 & 18.09 & 48.92 \\
    \hline
    TARN-VIST  & \textbf{30.47}      &\textbf{18.51}       & \textbf{49.43} \\
    \hline
    \end{tabular}%
    \caption{\label{topic-vist-table2} Quantitative results on the VIST dataset for semantic understanding evaluation metric. For all these metrics, higher score means better performance.}%
\end{table*}%

\textbf{Topic Consistency Rewards Design} \ For a given image sequence, we define that the extracted topic information from the perspective of vision and language are $topic_{v}$ and $topic_{l}$. For the story generated by the model, we use RAKE to extract its topic information $topic_{c}$, and then propose three reward functions $r_{bleu}$, $r_{topic-cv}$ and $r_{topic-cl}$ based on the text cosine similarity and topic consistency:

\begin{equation}
    r_{bleu}= \\ sentence \text{-} bleu(story_c,story_g)
    \label{reward-bleu}
\end{equation}

\begin{equation}
    r_{topic-cv}= cosine \text{-} {similarity}(topic_c,topic_{v})
    \label{reward-topic1}
\end{equation}

\begin{equation}
    r_{topic-cl}= cosine \text{-} {similarity}(topic_c,topic_{l})
    \label{reward_topic2}
\end{equation}

where $story\_c$ means the story generated by our model and $story\_g$ represents the corresponding ground-truth story. In summary, the total reward function is:

\begin{equation}
    r=\lambda\cdot r_{bleu} + \gamma \cdot r_{topic-cv}  + \eta \cdot r_{topic-cl}
    \label{reward-final}
\end{equation}

where $\lambda$, $\gamma$ and $\eta$ are the hyper-parameters controlling the proportion of each part.

\subsection{Training}

Reinforcement learning can learn a policy that encourages the model to focus more on those key aspects and then generate stories with more emotion and fluency through maximizing the given reward:

\begin{equation}
\label{eq10}\
J_{RL}({\beta})=\sum_{Y,V\in{D^{\prime}}}{E_{{y_{i}}\sim{}{\pi_{i}}}}{\left[\left(b-r\left({y_{i}}\right)\right)\log{\pi_{i}}\right]}
\end{equation}

\begin{equation}
\label{eq11}
r\left({y_{i}}\right)={\lambda}{r_{bleu}\left({y_{i}}\right)}+
{\gamma}{r_{topic-cv}\left({y_{i}}\right)}+{\eta}{r_{topic-cl}\left({y_{i}}\right)}\
\end{equation}

where $\pi$$\equiv$$p_{\boldsymbol{\theta}}\left(y_{i} \mid \mathbf{v}_{i}, \overline{\mathbf{v}}\right)$ is the policy and $b$ is the baseline that reduces the given reward variance. 

During training, we find that optimizing only with the reinforcement learning loss has the potential to increase expected rewards while sacrificing the quality of our generative model. To address this concern, we first train our model with maximum likelihood estimation (MLE) and then continue to train the model jointly with reinforcement loss ${Loss}_{R L}$ and MLE loss ${Loss}_{MLE}$. This training strategy can strike a balance between reward optimization and model fidelity, resulting in improved storytelling ability. MLE loss and the mixed training loss ${Loss}_{{mixed}}$ are defined as follows: 

\begin{equation}
\label{eq16}
Loss_{MLE}=\sum_{Y,V\in D^{’}} \sum_{i=1}^{N} \sum_{t=1}^{T} -log \left.P(W_{i,t}=Y_{i,t}\right)
\end{equation}

\begin{equation}
\label{eq14}
{Loss}_{{mixed}}=\omega {Loss}_{RL}+(1-\omega) {Loss}_{MLE}    
\end{equation}

Following \citet{2}, we set the $\omega$=0.5 in eq. \eqref{eq14} as it is a good trade-off between reinforcement learning loss and MLE loss. 

\section{Experiments}

\subsection{Dataset and Evaluation Metrics}

We train and evaluate our model on the VIST dataset~\cite{29}, which is the most widely used dataset for the visual storytelling task. The VIST dataset\footnote[1]{https://visionandlanguage.net/VIST/} includes 10,117 Flicker albums with 21,0819 unique images and it is split into training/validation/testing sets with 400,98/4,988/5,050 samples, respectively. 

During the whole experiment process, we utilize two groups of automatic metrics for the quantitative evaluation. One group consists of several surface-level-based similarity measures, including BLEU \cite{21}, METEOR \cite{22}, ROUGE \cite{23}, CIDEr \cite{24} and SPICE \cite{25}. In particular, BLEU is a classical metric that computes the geometric average of overlapping n-grams between the generated sentence and the reference sentence. Commonly used ones are BLEU-1, BLEU-2 and BLEU-4. METEOR calculates the sentence-level similarity scores based on the harmonic mean of uni-gram recall and precision. It is sensitive to the text length. ROUGE is a recall-based metric which captures the length of the most common sub-sequence between the generated story and the reference. CIDEr adopts higher order n-grams to account for fluency and uses term frequency-inverse document frequency (TF-IDF) weighting for each n-gram to down-weigh commonly occurring ones. SPICE is an automated caption evaluation metric that defined over scene graphs. It can effectively capture and reflect human judgments over model-generated captions.

\begin{table*}[!htbp]
  \centering
    \begin{tabular}{c|cccc}
    \hline
    Model & BLEU-4 & METEOR & CIDEr & SPICE \\
    \hline
    Baseline    & 12.40  & 33.90  & 8.60  & 8.30  \\
    Baseline+$r_{bleu}$ & 12.82  & 35.36  & 11.58  & 10.84  \\
    Baseline+$r_{bleu}$+$r_{topic-cv}$    & 13.10  & 35.62  & 11.70  & \textbf{11.37}  \\
    Baseline+$r_{bleu}$+$r_{topic-cl}$    & 13.44  & 35.80  & 10.99  & 11.36  \\
    \hline
    TARN-VIST  & \textbf{13.46}  & \textbf{35.88}  & \textbf{12.07}  & 11.25  \\
    \hline
    \end{tabular}%
    \caption{\label{topic-vist-ablation1} Ablation experiment results on different combinations of the reward functions. Note that here our basic model is ReCo-RL.}
\end{table*}%

\begin{table*}[!htbp]
  \centering
    \begin{tabular}{cc|cccccccc}
    \hline
    $\gamma$     & $\eta$     & BLEU-1 & BLEU-2 & BLEU-3 & BLEU-4 & ROUGE & METEOR & CIDEr & SPICE \\
    \hline
    0.3   & 0.7   & 68.43  & 42.48  & 23.51  & 12.85  & 29.14  & 35.52  & 11.23  & 10.76  \\
    0.4   & 0.6   & 68.38  & 42.55  & 23.56  & 12.88  & 29.21  & 35.62  & 11.69  & 10.85  \\
    0.5   & 0.5   & \textbf{69.01} & \textbf{43.56} & \textbf{24.27} & \textbf{13.46} & \textbf{29.28} & \textbf{35.88}  & \textbf{12.07} & \textbf{11.36} \\
    0.6   & 0.4   & 68.58  & 42.79  & 23.77  & 13.09  & 29.24  & 35.63  & 11.91  & 10.94  \\
    0.7   & 0.3   & 68.24  & 42.46  & 23.57  & 12.92  & 29.12  & 35.53  & 11.44  & 10.78  \\
    0.8   & 0.2   & 68.24  & 42.58  & 23.66  & 12.90  & 29.24  & 35.48  & 11.69  & 10.77  \\
    \hline
    \end{tabular}%
    \caption{\label{topic-vist-ablation2}Ablation experiment results of TARN-VIST with different $\gamma$ and $\theta$.}
\end{table*}%

\begin{table*}[!htbp]
  \centering
    \begin{tabular}{c|ccc|ccc}
    \hline
     Aspect & KE-VIST & TARN-VIST  & Tie   & PR-VIST & TARN-VIST  & Tie \\
    \hline
    Relevance & 19\%  & \textbf{67\%}  & 14\%  & 30\%  & \textbf{60\%}  & 10\% \\
    Coherence & 22\%  & \textbf{50\%}  & 28\%  & 18\%  & \textbf{62\%}  & 20\% \\
    Information Richness & 28\%  & \textbf{64\%}  & 8\%  & 33\%  & \textbf{50\%}  & 17\% \\
    \hline
    \end{tabular}%
    \caption{\label{topic-vist-table3}Human Pairwise Evaluation between TARN-VIST and other models. For each pairwise comparison, each of the three columns stands for the percentage of volunteers that prefer this story to the other one, and consider both stories are of equal quality.}
\end{table*}%

What’s more, since previous researchers \cite{5} have pointed out that current surface-level automatic metrics may correlate poorly with human judgments, we also use BERTScore \cite{26}, BARTScore \cite{27} and BLEURT \cite{28} in the expriments. Among them, BERTScore measures the BERT embedding similarity between each token as a more semantic and robust measure rather than common string-match-based similarity measure. BARTScore is an unsupervised metric that does not require human judgment. It can better evaluate the generated text from various aspects. BLEURT is a transfer learning based metric for natural language generation. It excels in capturing significant semantic similarities between sentences.

\begin{figure*}[t]
    \centering
\includegraphics[width=0.93\textwidth,height=0.51\textwidth]{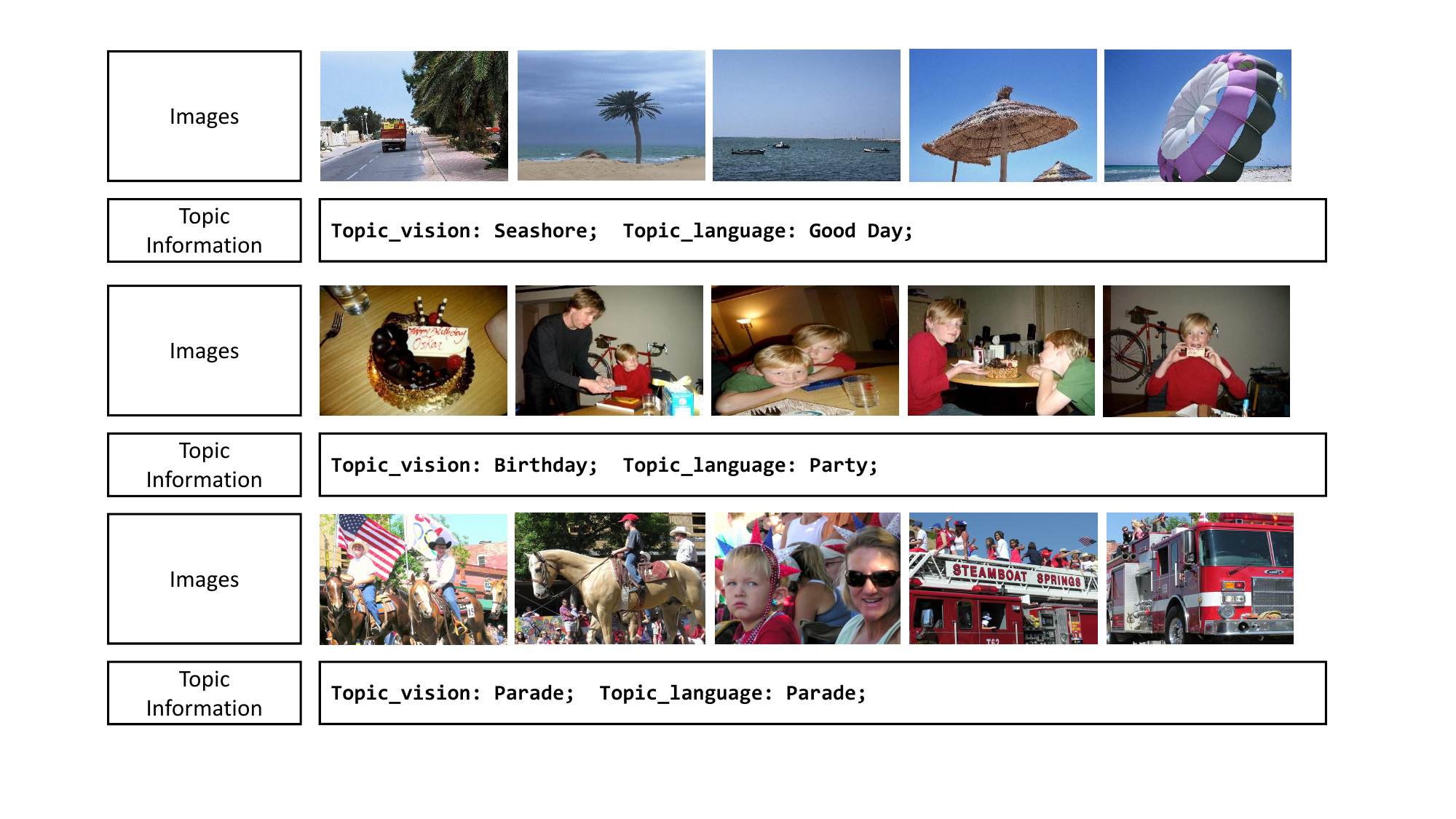}
    \caption{Examples of extracted topic information}
    \label{figure2}
\end{figure*}

\begin{figure*}[!htbp]
    \centering
 \includegraphics[width=0.93\textwidth,height=0.41\textwidth]{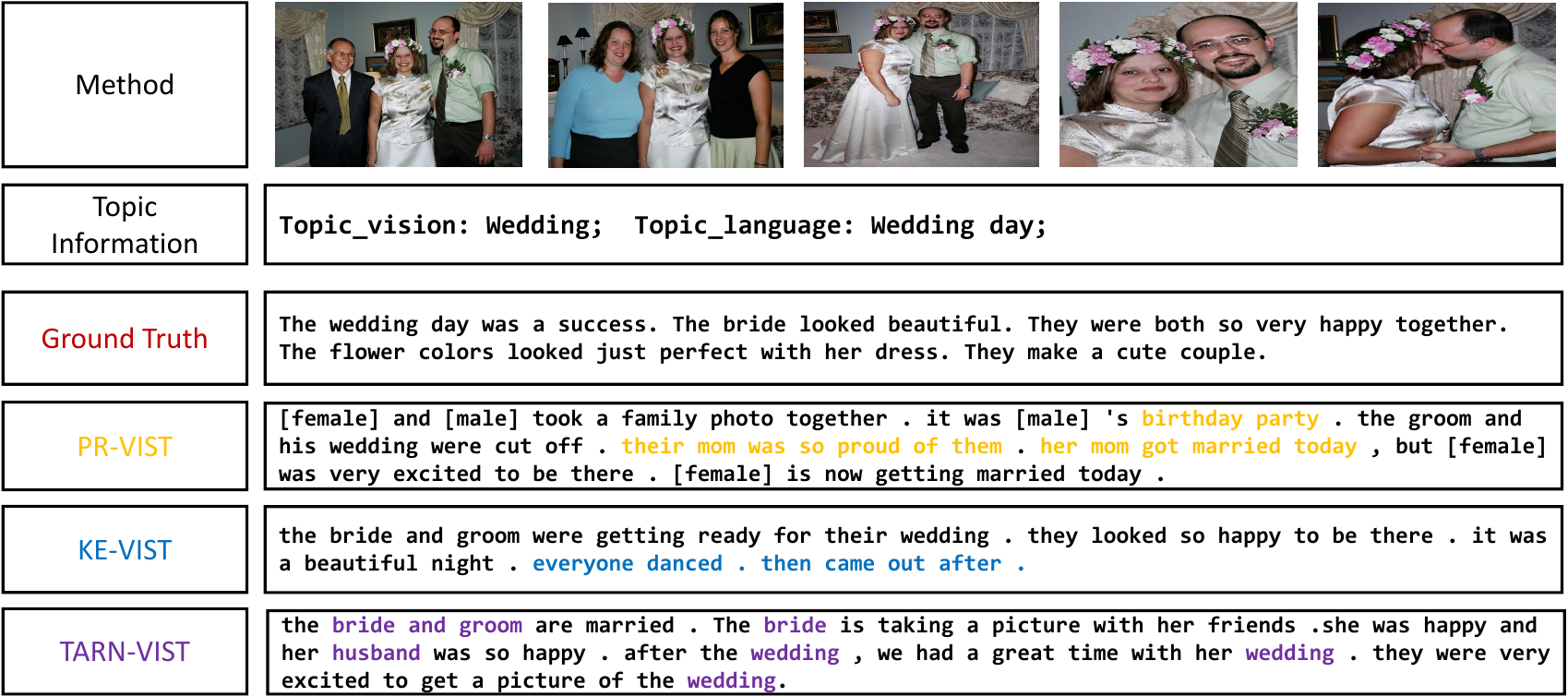}
    \caption{Example story generated from TARN-VIST and several competitive baselines.}
    \label{figure3}
\end{figure*}

\subsection{Implementation Details}

We implement our framework in PyTorch~\cite{44}. In the phase of encoding, we leverage the official pre-trained Resnet-152 model\footnote[1]{https://huggingface.co/microsoft/resnet-152} to extract the deeper image features. Additionally, in the phase of decoding, the hidden size of manager LSTM and worker LSTM are both set to 512. When calculating the cosine similarity, we employ the Bert-base model from HuggingFace\footnote[2]{https://huggingface.co/google-bert/bert-base-uncased}. Furthermore, during the model training with MLE loss, we set learning rate to 0.0001 and the dropout rate to 0.6. Our experiments are conducted on GeForce RTX 3090 GPU with a batch size of 128.

\subsection{Quantitative Evaluation}

We compare our model with the following baselines: (1) {Seq2seq} \cite{29}, a classical end-to-end sequence learning approach; (2) H-Attn-Rank \cite{32}, a model composed of three hierarchically-attentive recurrent neural nets (RNNs); (3) {BARNN} \cite{45}, an attention-based RNN with a skip gated recurrent unit (GRU) which leverages the semantic relation between photo streams and stories; (4) {Show, Reward and Tell (SRT)} \cite{42}, a hierarchical generative model with reinforcement learning and adversarial training; (5) {XE-ss} and {AREL} \cite{5}, an adversarial reward learning framework with imitation learning and GAN; (6) {HPSR} \cite{4}, a novel model with a hierarchical photo-scene encoder and a reconstructor; (7) {HSRL} \cite{20}, a structured reinforcement learning approach with hierarchical decoder; (8) {SGVST} \cite{13}, a framework based on scene graphs with GCN and TCN; (9) {ReCo-RL} \cite{2}, a reinforcement learning algorithm while the reward is relevance, coherence and expressiveness; (10) {INet} \cite{30}, a hide-and-tell training scheme; (11) {KE-VIST} \cite{8}, a three-stage generation framework; (12) {IRW} \cite{6}, an imagine-reason-write generation framework; (13) {CKAKS} \cite{7}, an extract-enrich-generate framework; (14) {PR-VIST}~\cite{9}, a plot and rework framework; (15) {LMGT}~\cite{37}, a Transformer based framework for visual story generation; (16) {Sentistory} \cite{31}, a sentiment-aware generative model.

In table \eqref{topic-vist-tabel1}, we observe that our proposed model TARN-VIST achieves state-of-the-art on the BLEU-1, BLEU-2, METEOR, CIDEr and SPICE. Specifically, compared with the previous best model, TARN-VIST makes the improvement by 1.5\% on BLEU-1, 1.9\% on BLEU-2, 0.2\% on METEOR, 1.1\% on CIDEr and 2.4\% on SPICE. In addition, the performance of BLEU-4 and ROUGE is also very competitive. In table \eqref{topic-vist-table2}, we find that TARN-VIST achieves state-of-the-art on all the metrics. The reason for the excellent performance of TARN-VIST can be attributed to our proposed composite reward function, which can enable the model to effectively achieve topic alignment and generate high-quality stories.

\subsection{Ablation Study}

To analyze the effectiveness of our proposed reward functions, we carry out a number of ablation experiments on different combinations of the reward functions. The experimental results are shown in Table \eqref{topic-vist-ablation1}. Note that our basic model is ReCo-RL \cite{2}, which is an encoder-decoder framework trained through reinforcement learning. From the table, we can see that when BLEU is used as the reward function of reinforcement learning, the performance of the model has been improved to a certain extent compared with the baseline. For instance, the evaluation indicators of CIDEr and SPICE have increased by 2.98\% and 2.54\%, respectively. Besides, when using $r_{topic-cv}$ or $r_{topic-cl}$ alone, the model also has different degrees of improvement. In addition, when $r_{bleu}$ and $r_{topic-cv}$ are combined, the relative improvements of CIDEr and SPICE are 0.12\% and 0.53\%. Furthermore, when $r_{bleu}$, $r_{topic-cl}$ and $r_{topic-cv}$ are integrated together, the model performs best on most evaluation metrics.

In eq. \eqref{eq11}, $\lambda$, $\gamma$ and $\eta$ are three important hyper-parameters. Therefore, we also implement different combinations of the hyper-parameters. To be noticed, $\lambda$ is set to 0.5 by the control variable method and the sum of $\gamma$ and $\eta$ is set to $1$. We carry out several experiments on different value pairs of $\gamma$ and $\eta$. The experimental results are shown in Table \eqref{topic-vist-ablation2}. It can be obviously seen that when $\gamma$ and $\eta$ are all equal to 0.5, the model can better balance the effects of the three reward functions, thus making the model reach a local optimal result.

\subsection{Human Evaluation}

Previous visual storytelling work has highlighted that current string-match-based automatic metrics are not a perfect way to evaluate the model's performance~\cite{5}. Therefore, we further conduct pairwise human evaluation to assess the subjective quality of our model. In particular, we randomly selected 100 generated sample stories from the test set and invited 20 well-educated volunteers to evaluate results based on the story relevance, the story coherence, and the story information richness. All the evaluators have received college degree or above. Meanwhile, the choices are shuffled before evaluation to avoid any bias. Our model is compared with KE-VIST \cite{8} and PR-VIST \cite{9}. For each sample pair, given a photo stream and the two stories generated by two models, the volunteers are asked to perform pairwise evaluation based on the relevance, coherence, and information richness. Among them, relevance assesses how the story accurately describes what is happening in the image sequence. Coherence evaluates whether the story is semantically coherent with other sentences. Information richness assesses whether the information is fully expressed in the story. To be more fair, we also a provide neutral option for cases that volunteers consider the two stories to be equally good on one particular criterion. The experimental results are displayed in Table \eqref{topic-vist-table3}. From the table, we can observe that our proposed model significantly outperforms other models in term of these three metrics. 

\subsection{Qualitative Evaluation}

In order to provide a more intuitively analysis on the rationality of our proposed topic information extraction method, we randomly choose several image sequences, and the topic information $topic_{v}$ and $topic_{l}$ extracted from the visual and linguistic perspectives are also displayed. The experimental results are shown in Fig. \eqref{figure2}. It can be found that both the topic information can well summarize the content of the image sequence. Take the first set of image sequences as an illustration, the topic information extracted from the visual perspective is "Seashore" and the topic information extracted from the linguistic perspective is "Good Day". We infer that when extracting the topic information from the visual perspective, multiple beach-related vocabulary would be extracted when using the image recognition module to extract the concept of the input image sequence, so the final topic information is "Seashore". Meanwhile, when extracting topic vocabulary from the linguistic perspective, since the story in the dataset describes the picture from the protagonist's point of view, the content is probably "spent a good day at the beach" and other information. Therefore, the extracted topic information is "Good Day".

In addition, to qualitatively evaluate the TARN-VIST, we also compares the sample stories generated by this model with the other two models in a specific image sequence. The results are exhibited in Fig. \eqref{figure3}. We can see that the image sequence topics extracted by our model are closely related to weddings. For the stories generated by the PR-VIST model, "birthday party", "their room was so proud of them" and "her mom got married today" are less pertinent to the topic of the input image sequence. For the stories generated by the KE-VIST model, the two short sentences "everyone danced" and "then came out after" have serious expression problems such as grammatical errors and low coherence. On the contrary, the stories generated by the TARN-VIST are more relevant than those generated by the other two models. For instance, in each sentence, there are words such as "bride", "groom" and "wedding". The presence of a story brings the tone of the story and the topic information closer together so that the stories generated by TARN-VIST are more coherent.

\section{Conclusion and Future Work}

In this paper, we propose TARN-VIST, an innovative topic aware reinforcement network for visual storytelling to generate more cogent story. We first utilize CLIP and RAKE to mine the topic information of the story from both visual and linguistic perspectives. Subsequently, we employ the reinforcement learning and design topic consistency rewards to refine the generation process. Extensive experimental results on the benchmark dataset demonstrate that our model outperforms most competitive baselines across a number of evaluation metrics.

In future work, we will explore grammar and discourse structure in visual storytelling tasks, as they play a key role in the accuracy, coherence, and readability of the generated stories. Besides, it is also interesting and practical to further analyse linguistic style to improve the quality and diversity of generated stories.

\section{Acknowledgements}

This work was partially supported by Postgraduate Research \& Practice Innovation Program of Jiangsu Province, National Natural Science Foundation of China (NSFC Grant No. 61773272, 61272258, 61301299, 615-72085, 61170124, 61272005, 62376041), Provincial Natural Science Foundation of Jiangsu, China (Grant No. BK20151254, BK201512-60), Science and Education Innovation based Cloud Data fusion Foundation of Science and Technology Development Center of Education Ministry, China (2017B03112), Six talent peaks Project in
Jiangsu Province, China (DZXX-027), Key Laboratory of Symbolic Computation and Knowledge Engineering of Ministry of Education, Jilin University, China (Grant No.93K172016K08), and Collaborative Innovation Center of Novel Software Technology and Industrialization, China.

\clearpage

\nocite{*}
\section{Bibliographical References}\label{sec:reference}

\bibliographystyle{lrec-coling2024-natbib}
\bibliography{lrec-coling2024-example}

\begin{thebibliography}{44}
\expandafter\ifx\csname natexlab\endcsname\relax\def\natexlab#1{#1}\fi

\bibitem[{Anderson et~al.(2016)Anderson, Fernando, Johnson, and Gould}]{25}
Peter Anderson, Basura Fernando, Mark Johnson, and Stephen Gould. 2016.
\newblock {SPICE:} semantic propositional image caption evaluation.
\newblock In \emph{Computer Vision - {ECCV} 14th European Conference}, pages 382--398, Amsterdam, The Netherlands. Springer.

\bibitem[{Banerjee and Lavie(2005)}]{22}
Satanjeev Banerjee and Alon Lavie. 2005.
\newblock {METEOR:} an automatic metric for {MT} evaluation with improved correlation with human judgments.
\newblock In \emph{Proceedings of the Workshop on Intrinsic and Extrinsic Evaluation Measures for Machine Translation and/or Summarization@ACL 2005, Ann Arbor, Michigan, USA, June 29, 2005}, pages 65--72, Ann Arbor, Michigan, USA. Association for Computational Linguistics.

\bibitem[{Braude et~al.(2022)Braude, Schwartz, Schwing, and Shamir}]{14}
Tom Braude, Idan Schwartz, Alexander~G. Schwing, and Ariel Shamir. 2022.
\newblock Ordered attention for coherent visual storytelling.
\newblock In \emph{{MM} '22: The 30th {ACM} International Conference on Multimedia}, pages 3310--3318, Lisboa, Portugal. {ACM}.

\bibitem[{Chen et~al.(2021)Chen, Huang, Takamura, and Nakayama}]{7}
Hong Chen, Yifei Huang, Hiroya Takamura, and Hideki Nakayama. 2021.
\newblock Commonsense knowledge aware concept selection for diverse and informative visual storytelling.
\newblock In \emph{Thirty-Fifth {AAAI} Conference on Artificial Intelligence}, pages 999--1008, Virtual Event. {AAAI} Press.

\bibitem[{Chen et~al.(2022)Chen, Liu, and Niu}]{31}
Wei Chen, Xuefeng Liu, and Jianwei Niu. 2022.
\newblock Sentistory: {A} multi-layered sentiment-aware generative model for visual storytelling.
\newblock \emph{{IEEE} Trans. Circuits Syst. Video Technol.}, 32(11):8051--8064.

\bibitem[{Chen et~al.(2018)Chen, Wang, Wang, and Wang}]{5}
Wenhu Chen, Xin Wang, Yuan{-}Fang Wang, and William~Yang Wang. 2018.
\newblock No metrics are perfect: Adversarial reward learning for visual storytelling.
\newblock In \emph{Proceedings of the 56th Annual Meeting of the Association for Computational Linguistics, {ACL}, Volume 1: Long Papers}, pages 899--909, Melbourne, Australia. Association for Computational Linguistics.

\bibitem[{Chu et~al.(2021)Chu, Hsu, Huang, and Ku}]{9}
Yun{-}Wei Chu, Chi{-}Yang Hsu, Ting{-}Hao~Kenneth Huang, and Lun{-}Wei Ku. 2021.
\newblock Plot and rework: Modeling storylines for visual storytelling.
\newblock \emph{CoRR}, abs/2105.06950.

\bibitem[{Fan et~al.(2021)Fan, Wang, Gu, and Liu}]{1}
Ruichao Fan, Hanli Wang, Jinjing Gu, and Xianhui Liu. 2021.
\newblock Visual storytelling with hierarchical {BERT} semantic guidance.
\newblock In \emph{MMAsia '21: {ACM} Multimedia Asia}, pages 24:1--24:7, Gold Coast, Australia. {ACM}.

\bibitem[{Gu et~al.(2023)Gu, Wang, and Fan}]{41}
Jinjing Gu, Hanli Wang, and Ruichao Fan. 2023.
\newblock Coherent visual storytelling via parallel top-down visual and topic attention.
\newblock \emph{{IEEE} Trans. Circuits Syst. Video Technol.}, 33(1):257--268.

\bibitem[{He et~al.(2016)He, Zhang, Ren, and Sun}]{11}
Kaiming He, Xiangyu Zhang, Shaoqing Ren, and Jian Sun. 2016.
\newblock Deep residual learning for image recognition.
\newblock In \emph{{IEEE} Conference on Computer Vision and Pattern Recognition, {CVPR}}, pages 770--778, Las Vegas, NV, USA. {IEEE} Computer Society.

\bibitem[{Hsu et~al.(2020)Hsu, Chen, Hsu, Li, Lin, Huang, and Ku}]{8}
Chao{-}Chun Hsu, Zi{-}Yuan Chen, Chi{-}Yang Hsu, Chih{-}Chia Li, Tzu{-}Yuan Lin, Ting{-}Hao~Kenneth Huang, and Lun{-}Wei Ku. 2020.
\newblock Knowledge-enriched visual storytelling.
\newblock In \emph{The Thirty-Fourth {AAAI} Conference on Artificial Intelligence}, pages 7952--7960, Honolulu, Hawaii, USA. {AAAI} Press.

\bibitem[{Hu et~al.(2020)Hu, Cheng, Gan, Liu, Gao, and Neubig}]{2}
Junjie Hu, Yu~Cheng, Zhe Gan, Jingjing Liu, Jianfeng Gao, and Graham Neubig. 2020.
\newblock What makes {A} good story? designing composite rewards for visual storytelling.
\newblock In \emph{The Thirty-Fourth {AAAI} Conference on Artificial Intelligence}, pages 7969--7976, New York, NY, USA. {AAAI} Press.

\bibitem[{Huang et~al.(2019)Huang, Gan, Celikyilmaz, Wu, Wang, and He}]{20}
Qiuyuan Huang, Zhe Gan, Asli Celikyilmaz, Dapeng~Oliver Wu, Jianfeng Wang, and Xiaodong He. 2019.
\newblock Hierarchically structured reinforcement learning for topically coherent visual story generation.
\newblock In \emph{The Thirty-Third {AAAI} Conference on Artificial Intelligence}, pages 8465--8472, Honolulu, Hawaii, USA. {AAAI} Press.

\bibitem[{Huang et~al.(2016)Huang, Ferraro, Mostafazadeh, Misra, Agrawal, Devlin, Girshick, He, Kohli, Batra, Zitnick, Parikh, Vanderwende, Galley, and Mitchell}]{29}
Ting{-}Hao~(Kenneth) Huang, Francis Ferraro, Nasrin Mostafazadeh, Ishan Misra, Aishwarya Agrawal, Jacob Devlin, Ross~B. Girshick, Xiaodong He, Pushmeet Kohli, Dhruv Batra, C.~Lawrence Zitnick, Devi Parikh, Lucy Vanderwende, Michel Galley, and Margaret Mitchell. 2016.
\newblock Visual storytelling.
\newblock In \emph{{NAACL} {HLT} The 2016 Conference of the North American Chapter of the Association for Computational Linguistics: Human Language Technologies}, pages 1233--1239, San Diego California, USA. The Association for Computational Linguistics.

\bibitem[{Jung et~al.(2020)Jung, Kim, Woo, Kim, Kim, and Kweon}]{30}
Yunjae Jung, Dahun Kim, Sanghyun Woo, Kyungsu Kim, Sungjin Kim, and In~So Kweon. 2020.
\newblock Hide-and-tell: Learning to bridge photo streams for visual storytelling.
\newblock In \emph{The Thirty-Fourth {AAAI} Conference on Artificial Intelligence, {AAAI}}, pages 11213--11220, New York, NY, USA. {AAAI} Press.

\bibitem[{Kim et~al.(2018)Kim, Heo, Son, Park, and Zhang}]{3}
Taehyeong Kim, Min{-}Oh Heo, Seonil Son, Kyoung{-}Wha Park, and Byoung{-}Tak Zhang. 2018.
\newblock {GLAC} net: Glocal attention cascading networks for multi-image cued story generation.
\newblock \emph{CoRR}, abs/1805.10973.

\bibitem[{Kiran et~al.(2022)Kiran, Sobh, Talpaert, Mannion, Sallab, Yogamani, and P{\'{e}}rez}]{16}
B.~Ravi Kiran, Ibrahim Sobh, Victor Talpaert, Patrick Mannion, Ahmad A.~Al Sallab, Senthil~Kumar Yogamani, and Patrick P{\'{e}}rez. 2022.
\newblock Deep reinforcement learning for autonomous driving: {A} survey.
\newblock \emph{{IEEE} Trans. Intell. Transp. Syst.}, 23(6):4909--4926.

\bibitem[{Li et~al.(2020)Li, Tang, Li, Xiao, Wu, Pu, and Zhuang}]{15}
Jiacheng Li, Siliang Tang, Juncheng Li, Jun Xiao, Fei Wu, Shiliang Pu, and Yueting Zhuang. 2020.
\newblock Topic adaptation and prototype encoding for few-shot visual storytelling.
\newblock In \emph{{MM} '20: The 28th {ACM} International Conference on Multimedia}, pages 4208--4216, Virtual Event / Seattle, WA. {ACM}.

\bibitem[{Li et~al.(2023)Li, Liu, and Ji}]{40}
Xin Li, Chunping Liu, and Yi~Ji. 2023.
\newblock Associative learning network for coherent visual storytelling.
\newblock In \emph{ICASSP 2023 - 2023 IEEE International Conference on Acoustics, Speech and Signal Processing (ICASSP)}, page~5, RHODES ISLAND, GREECE. {IEEE}.

\bibitem[{Lin(2004)}]{23}
Chin-Yew Lin. 2004.
\newblock Rouge: A package for automatic evaluation of summaries.
\newblock In \emph{Proceedings of the 42nd Annual Meeting of the Association for Computational Linguistics}, page 74–81, Barcelona, Spain. {ACL}.

\bibitem[{Liu et~al.(2017)Liu, Fu, Mei, and Chen}]{45}
Yu~Liu, Jianlong Fu, Tao Mei, and Changen Chen. 2017.
\newblock Let your photos talk: Generating narrative paragraph for photo stream via bidirectional attention recurrent neural networks.
\newblock In \emph{Proceedings of the Thirty-First {AAAI} Conference on Artificial Intelligence, (AAAI-17)}, page~8, San Francisco, California USA. {AAAI} Press.

\bibitem[{Mnih et~al.(2013)Mnih, Kavukcuoglu, Silver, Graves, Antonoglou, Wierstra, and Riedmiller}]{17}
Volodymyr Mnih, Koray Kavukcuoglu, David Silver, Alex Graves, Ioannis Antonoglou, Daan Wierstra, and Martin~A. Riedmiller. 2013.
\newblock Playing atari with deep reinforcement learning.
\newblock \emph{CoRR}, abs/1312.5602.

\bibitem[{Nie et~al.(2023)Nie, Chen, and Wang}]{33}
Mingshuo Nie, Dongming Chen, and Dongqi Wang. 2023.
\newblock Reinforcement learning on graphs: {A} survey.
\newblock \emph{{IEEE} Trans. Emerg. Top. Comput. Intell.}, 7(4):1065--1082.

\bibitem[{Orr and Dutta(2023)}]{35}
James Orr and Ayan Dutta. 2023.
\newblock Multi-agent deep reinforcement learning for multi-robot applications: {A} survey.
\newblock \emph{Sensors}, 23(7):3625.

\bibitem[{Papineni et~al.(2002)Papineni, Roukos, Ward, and Zhu}]{21}
Kishore Papineni, Salim Roukos, Todd Ward, and Wei{-}Jing Zhu. 2002.
\newblock Bleu: a method for automatic evaluation of machine translation.
\newblock In \emph{Proceedings of the 40th Annual Meeting of the Association for Computational Linguistics}, pages 311--318, Philadelphia, PA, {USA}. {ACL}.

\bibitem[{Paszke et~al.(2019)Paszke, Gross, Massa, Lerer, Bradbury, Chanan, Killeen, Lin, Gimelshein, Antiga, Desmaison, Kopf, Yang, DeVito, Raison, Tejani, Chilamkurthy, Steiner, Fang, Bai, and Chintala}]{44}
Adam Paszke, Sam Gross, Francisco Massa, Adam Lerer, James Bradbury, Gregory Chanan, Trevor Killeen, Zeming Lin, Natalia Gimelshein, Luca Antiga, Alban Desmaison, Andreas Kopf, Edward Yang, Zachary DeVito, Martin Raison, Alykhan Tejani, Sasank Chilamkurthy, Benoit Steiner, Lu~Fang, Junjie Bai, and Soumith Chintala. 2019.
\newblock Pytorch: An imperative style, high-performance deep learning library.
\newblock In \emph{Advances in Neural Information Processing Systems}, page~12, Vancouver, BC, Canada. Curran Associates, Inc.

\bibitem[{Perolat et~al.(2022)Perolat, Vylder, Hennes, Tarassov, Strub, de~Boer, Muller, Connor, Burch, Anthony, McAleer, Elie, Cen, Wang, Gruslys, Malysheva, Khan, Ozair, Timbers, Pohlen, Eccles, Rowland, Lanctot, Lespiau, Piot, Omidshafiei, Lockhart, Sifre, Beauguerlange, Munos, Silver, Singh, Hassabi, and Tuyls}]{43}
Julien Perolat, Bart~De Vylder, Daniel Hennes, Eugene Tarassov, Florian Strub, Vincent de~Boer, Paul Muller, Jerome~T. Connor, Neil Burch, Thomas Anthony, Stephen McAleer, Romuald Elie, Sarah~H. Cen, Zhe Wang, Audrunas Gruslys, Aleksandra Malysheva, Mina Khan, Sherjil Ozair, Finbarr Timbers, Toby Pohlen, Tom Eccles, Mark Rowland, Marc Lanctot, Jean-Baptiste Lespiau, Bilal Piot, Shayegan Omidshafiei, Edward Lockhart, Laurent Sifre, Nathalie Beauguerlange, Remi Munos, David Silver, Satinder Singh, Demis Hassabi, and Karl Tuyls. 2022.
\newblock Mastering the game of stratego with model-free multiagent reinforcement learning.
\newblock \emph{Science}, 378(6623):990--996.

\bibitem[{Plaat et~al.(2023)Plaat, Kosters, and Preuss}]{34}
Aske Plaat, Walter~A. Kosters, and Mike Preuss. 2023.
\newblock High-accuracy model-based reinforcement learning, a survey.
\newblock \emph{Artif. Intell. Rev.}, 56(9):9541--9573.

\bibitem[{Qi et~al.(2021)Qi, Qin, Huang, Shen, Yang, and Luo}]{37}
Mengshi Qi, Jie Qin, Di~Huang, Zhiqiang Shen, Yi~Yang, and Jiebo Luo. 2021.
\newblock Latent memory-augmented graph transformer for visual storytelling.
\newblock In \emph{{MM} '21: {ACM} Multimedia Conference}, pages 4892--4901, Virtual Event. {ACM}.

\bibitem[{Radford et~al.(2021)Radford, Kim, Hallacy, Ramesh, Goh, Agarwal, Sastry, Askell, Mishkin, Clark, Krueger, and Sutskever}]{18}
Alec Radford, Jong~Wook Kim, Chris Hallacy, Aditya Ramesh, Gabriel Goh, Sandhini Agarwal, Girish Sastry, Amanda Askell, Pamela Mishkin, Jack Clark, Gretchen Krueger, and Ilya Sutskever. 2021.
\newblock Learning transferable visual models from natural language supervision.
\newblock In \emph{Proceedings of the 38th International Conference on Machine Learning, {ICML}}, pages 8748--8763, Virtual Event. {PMLR}.

\bibitem[{Rose et~al.(2010)Rose, Engel, Cramer, and Cowley}]{19}
Stuart~J Rose, David~W Engel, Nicholas~O Cramer, and Wendy~E Cowley. 2010.
\newblock Automatic keyword extraction from individual documents.

\bibitem[{Sellam et~al.(2020)Sellam, Das, and Parikh}]{28}
Thibault Sellam, Dipanjan Das, and Ankur~P. Parikh. 2020.
\newblock {BLEURT:} learning robust metrics for text generation.
\newblock In \emph{Proceedings of the 58th Annual Meeting of the Association for Computational Linguistics, {ACL}}, pages 7881--7892, Online event. Association for Computational Linguistics.

\bibitem[{Simonyan and Zisserman(2015)}]{10}
Karen Simonyan and Andrew Zisserman. 2015.
\newblock Very deep convolutional networks for large-scale image recognition.
\newblock In \emph{3rd International Conference on Learning Representations, {ICLR}}, San Diego, CA, USA.

\bibitem[{Vedantam et~al.(2015)Vedantam, Zitnick, and Parikh}]{24}
Ramakrishna Vedantam, C.~Lawrence Zitnick, and Devi Parikh. 2015.
\newblock Cider: Consensus-based image description evaluation.
\newblock In \emph{{IEEE} Conference on Computer Vision and Pattern Recognition, {CVPR}}, pages 4566--4575, Boston, MA, USA. {IEEE} Computer Society.

\bibitem[{Wang et~al.(2019)Wang, Ma, Zhang, Jiang, and Zhang}]{4}
Bairui Wang, Lin Ma, Wei Zhang, Wenhao Jiang, and Feng Zhang. 2019.
\newblock Hierarchical photo-scene encoder for album storytelling.
\newblock In \emph{The Thirty-Third {AAAI} Conference on Artificial Intelligence}, pages 8909--8916, Honolulu, Hawaii, USA. {AAAI} Press.

\bibitem[{Wang et~al.(2018)Wang, Fu, Tang, Li, and Mei}]{42}
Jing Wang, Jianlong Fu, Jinhui Tang, Zechao Li, and Tao Mei. 2018.
\newblock Show, reward and tell: Automatic generation of narrative paragraph from photo stream by adversarial training.
\newblock In \emph{Proceedings of the Thirty-Second {AAAI} Conference on Artificial Intelligence, (AAAI-18)}, page~8, New Orleans, Louisiana, USA. {AAAI} Press.

\bibitem[{Wang et~al.(2020)Wang, Wei, Li, Zhang, and Huang}]{13}
Ruize Wang, Zhongyu Wei, Piji Li, Qi~Zhang, and Xuanjing Huang. 2020.
\newblock Storytelling from an image stream using scene graphs.
\newblock In \emph{The Thirty-Fourth {AAAI} Conference on Artificial Intelligence}, pages 9185--9192, New York, NY, USA. {AAAI} Press.

\bibitem[{Xu et~al.(2021)Xu, Yang, Li, Shen, Ao, and Xu}]{6}
Chunpu Xu, Min Yang, Chengming Li, Ying Shen, Xiang Ao, and Ruifeng Xu. 2021.
\newblock Imagine, reason and write: Visual storytelling with graph knowledge and relational reasoning.
\newblock In \emph{Thirty-Fifth {AAAI} Conference on Artificial Intelligence}, pages 3022--3029, Virtual Event. {AAAI} Press.

\bibitem[{Yang and Jin(2023)}]{38}
Dingyi Yang and Qin Jin. 2023.
\newblock Attractive storyteller: Stylized visual storytelling with unpaired text.
\newblock In \emph{Proceedings of the 61st Annual Meeting of the Association for Computational Linguistics (Volume 1: Long Papers)}, pages 11053--11066, Toronto, Canada. Association for Computational Linguistics.

\bibitem[{Yang et~al.(2019)Yang, Luo, Chen, Li, Yin, He, and Sun}]{12}
Pengcheng Yang, Fuli Luo, Peng Chen, Lei Li, Zhiyi Yin, Xiaodong He, and Xu~Sun. 2019.
\newblock Knowledgeable storyteller: {A} commonsense-driven generative model for visual storytelling.
\newblock In \emph{Proceedings of the Twenty-Eighth International Joint Conference on Artificial Intelligence, {IJCAI}}, pages 5356--5362, Macao, China. ijcai.org.

\bibitem[{Yu et~al.(2023)Yu, Liu, Nemati, and Yin}]{36}
Chao Yu, Jiming Liu, Shamim Nemati, and Guosheng Yin. 2023.
\newblock Reinforcement learning in healthcare: {A} survey.
\newblock \emph{{ACM} Comput. Surv.}, 55(2):5:1--5:36.

\bibitem[{Yu et~al.(2017)Yu, Bansal, and Berg}]{32}
Licheng Yu, Mohit Bansal, and Tamara~L. Berg. 2017.
\newblock Hierarchically-attentive {RNN} for album summarization and storytelling.
\newblock In \emph{Proceedings of the 2017 Conference on Empirical Methods in Natural Language Processing, {EMNLP}}, pages 966--971, Copenhagen, Denmark. The Association for Computational Linguistics.

\bibitem[{Yuan et~al.(2021)Yuan, Neubig, and Liu}]{27}
Weizhe Yuan, Graham Neubig, and Pengfei Liu. 2021.
\newblock Bartscore: Evaluating generated text as text generation.
\newblock In \emph{Advances in Neural Information Processing Systems 34: Annual Conference on Neural Information Processing Systems}, pages 27263--27277, virtual event. MIT Press.

\bibitem[{Zhang et~al.(2020)Zhang, Kishore, Wu, Weinberger, and Artzi}]{26}
Tianyi Zhang, Varsha Kishore, Felix Wu, Kilian~Q. Weinberger, and Yoav Artzi. 2020.
\newblock Bertscore: Evaluating text generation with {BERT}.
\newblock In \emph{8th International Conference on Learning Representations, {ICLR}}, Addis Ababa, Ethiopia. OpenReview.net.

\end{thebibliography}

\end{document}